\documentclass[runningheads]{llncs}
\usepackage[T1]{fontenc}
\usepackage{graphicx}
\pdfoutput=1 
\usepackage{enumerate}
\usepackage{times}
\usepackage{epsfig}
\usepackage{graphicx}
\usepackage{bm}
\usepackage{booktabs}
\usepackage{multirow}
\usepackage{amsmath}
\usepackage{amssymb}
\usepackage{mathtools}
\usepackage{adjustbox}
\usepackage{xcolor}
\usepackage{mathrsfs}

\newcommand{\Diff}[0]{{\rm Diff}}

\usepackage{enumitem}
\usepackage{svg} %
\setlist{nosep, leftmargin=14pt}
\usepackage{indentfirst}
\usepackage[breaklinks=true,bookmarks=false]{hyperref}
\usepackage{algorithm}  
\usepackage[noend]{algpseudocode}
\usepackage{algorithmicx,algorithm}
\usepackage{amsthm}
\theoremstyle{definition}
\usepackage{ulem}

\begin{document}
%

%%%%%%%%% TITLE
\title{NeurEPDiff: Neural Operators to Predict Geodesics in Deformation Spaces}
\titlerunning{NeurEPDiff: Neural Operators to Predict Geodesics in Deformation Spaces}
\author{Nian Wu  \inst{1,2}\and
Miaomiao Zhang\inst{2,3}}
\authorrunning{Wu and Zhang}
\institute{Computer Science, East China Normal University, Shanghai, China \and 
Electrical \& Computer Engineering, University of Virginia, VA,USA \and
Computer Science, University of Virginia,  VA, USA}

\maketitle              % typeset the header of the contr ibution
\begin{abstract}
This paper presents NeurEPDiff, a novel network to fast predict the geodesics in deformation spaces generated by a well known Euler-Poincaré differential equation (EPDiff). To achieve this, we develop a neural operator that for the first time learns the evolving trajectory of geodesic deformations parameterized in the tangent space of diffeomorphisms  (a.k.a velocity fields). In contrast to previous methods that purely fit the training images, our proposed NeurEPDiff learns a nonlinear mapping function between the time-dependent velocity fields. A composition of integral operators and smooth activation functions is formulated in each layer of NeurEPDiff to effectively approximate such mappings. The fact that NeurEPDiff is able to rapidly provide the numerical solution of EPDiff (given any initial condition) results in a significantly reduced computational cost of geodesic shooting of diffeomorphisms in a high-dimensional image space. Additionally, the properties of discretiztion/resolution-invariant of NeurEPDiff make its performance generalizable to multiple image resolutions after being trained offline. We demonstrate the effectiveness of NeurEPDiff in registering two image datasets: 2D synthetic data and 3D brain resonance imaging (MRI). The registration accuracy and computational efficiency are compared with the state-of-the-art diffeomophic registration algorithms with geodesic shooting. 
\end{abstract}

\section{Introduction}
Deformable image registration is a fundamental tool in medical image analysis, for example, computational anatomy and shape analysis~\cite{niethammer2011geodesic,zhang2016low}, statistical analysis of groupwise images~\cite{o2012unbiased,wang2022geo}, and template-based image segmentation~\cite{ashburner2005unified,Joshi2004}. Among many applications, it is ideal that the transformation is diffeomorphic (i.e., bijective, smooth, and invertible smooth) to maintain an intact topology of objects presented in images. A bountiful literature has studied various parameterizations of diffeomorphisms, including stationary velocity fields~\cite{arsigny2006log,ashburner2007fast}, B-spline free-form deformations~\cite{schnabel2001generic,shi2012registration}, and the large deformation diffeomorphic metric mapping (LDDMM)~\cite{beg2005computing,vialard2012}. In this paper, we focus on LDDMM, which provides a rigorous mathematical definition of distance metrics on the space of diffeomorphisms. Such distance metrics play an important role in deformation-based shape analysis of images, such as geometric shape regression~\cite{hong2017fast,niethammer2011geodesic}, longitudinal shape analysis~\cite{singh2013hierarchical}, and group shape comparisons~\cite{miller2004computational,qiu2008parallel}. 

Despite the advantages of LDDMM~\cite{beg2005computing}, the substantial time and computational cost involved in searching for the optimal diffeomorphism on high-dimensional image grids limits its applicability in real-world applications. To alleviate this problem, a geodesic shooting algorithm~\cite{vialard2012} was developed to reformulate the optimization of LDDMM (originally over a sequence of time-dependent diffeomorphisms) equivalently in the tangent space of diffeomorphisms at time point zero (a.k.a., initial velocity fields). The geodesic is then uniquely determined by integrating a given initial velocity field via the Euler-Poincaré differential equation (EPDiff)~\cite{arnold1966,miller2006}. This geodesic shooting algorithm demonstrated a faster convergence and avoided having to store the entire time-varying velocity fields from one iterative search to the next. A recent work FLASH has further reduced the computational cost of geodesic shooting by reparameterizing the velocity fields in a low-dimensional bandlimited space~\cite{zhang2015fast,zhang2017frequency}. The numerical solution to EPDiff was rapidly computed with much fewer parameters in a frequency domain. In spite of a significant speedup, it still takes FLASH minutes to compute when the dimension of the bandlimited space grows higher. 

In this paper, we present a novel method, NeurEPDiff, that can fast predict the geodesics in deformation spaces generated by EPDiff. Inspired by the recent achievements on learning-based partial/ordinary differential equation solvers~\cite{li2020fourier,lu2019deeponet}, we develop a neural operator to learn the evolving functions/mappings of geodesics in the space of diffeomorphisms parameterized by bandlimited initial velocity fields. 
In contrast to current deep learning-based approaches that are tied to the resolution and discretization of training images~\cite{yang2017quicksilver,rohe2017svf,balakrishnan2019voxelmorph,cao2015semi,Wang_2020_CVPR}, our proposed method shows consistent performances on different scales of image resolutions without the need of re-training. We also define a composition of integral operators and smooth activation functions in each layer of the NeurEPDiff network to effectively approximate the mapping functions of EPDiff over time. To summarize, our main contributions are threefold: 
\begin{enumerate}[label=(\roman*)]
    \item We are the first to develop a neural operator that fast predicts a numerical solution to the EPDiff~\cite{arnold1966,miller2006}, which is a critical step to generate geodesics of diffeomorphisms.

    \item A composition of integral operators and smooth activation functions is defined to approximate the mapping functions of EPDiff characterized in the low-dimensional bandlimited space. This makes the training and inference of our proposed NeurEPDiff much more efficient in high-dimensional image spaces. 

    \item The performance of NeurEPDiff is invariant to image resolutions. That is to say, once it is trained, this operator can be tested on multiple image resolutions without the need of being re-trained. 
\end{enumerate}

To demonstrate the effectiveness of our proposed model in diffeomorphic image registration, we run experiments on both 2D synthetic data and 3D real brain MRIs. We first show that a trained NeurEPDiff is able to consistently predict the diffeomorphic transformations at multiple different image resolutions. We then compare the registration accuracy (via registration-based image segmentation) and computational efficiency of NeurEPDiff vs. the state-of-the-art LDDMM with geodesic shooting~\cite{zhang2015fast,zhang2017frequency,vialard2012}. Experimental results show that our method outperforms the baselines in term of time consumption, while with a comparable registration accuracy. 

%The rest of the article is divided as follows. In Sect.~\ref{sec:background} we provide a brief overview of diffeomorphic image registration in the LDDMM~\cite{beg2005computing} paradigm with geodesic shooting~\cite{vialard2012,younes2009evolutions} and of finite dimensional Fourier representation of dieomorphic deformations, and the fundamental aspects of deep learning based PDE solver Fourier neural operator~\cite{li2020fourier}. In Sect. 3 we present our method for diffeomorphic registration. Results in real and simulated datasets are presented in Sect. 4. Finally, Sect. 5 presents discussion and some concluding remarks.

% \section{Background: Deformable Image Registration}
% 
\section{Background: Geodesics of Diffeomorphisms via EPDiff}\label{sec:background}
%\subsection{Diffeomorphic Image Registration}
\label{sec:backgroundlddmm}
In this section, we first briefly review the basic concepts of {\em geodesic shooting}, which generates geodesics of diffeomorphisms via EPDiff~\cite{vialard2012,younes2009evolutions}. We then show how to optimize the problem of diffeomorphic image registration in the setting of LDDMM~\cite{beg2005computing} with geodesic shooting~\cite{vialard2012,younes2009evolutions}. 

\paragraph{\bf Geodesic shooting via EPDiff.} Let $\Diff^\infty(\Omega)$ denote the space of smooth
diffeomorphisms on an image domain $\Omega$. The tangent space of diffeomorphisms is the
space $V = \mathfrak{X}^\infty(T\Omega)$ of smooth vector fields on $\Omega$. Consider a time-varying velocity field, $\{v_t\} : [0,\tau] \rightarrow V$, we can generate diffeomorphisms $\{\phi_{t}\}$ as a solution to the equation
\begin{equation}
\label{eq:phi_v}
\frac{d \phi_t}{dt} = v_t(\phi_t), \, t \in [0, \tau].
\end{equation}

The geodesic shooting algorithm~\cite{Vialard2011DiffeomorphicAE} states that a geodesic path of the diffeomorphisms (in Eq.~\eqref{eq:phi_v}) is uniquely determined by integrating a well-studied EPDiff~\cite{arnold1966,miller2006} with an initial condition. That is, given an initial velocity, $v_0 \in V$, at $t
= 0$, a geodesic path $t \mapsto \phi_t \in \Diff^\infty(\Omega)$ in the space of diffeomorphisms can be computed by forward shooting the EPDiff equation
\begin{equation}
\label{eq:epdiff}
    \frac{\partial v_t}{\partial t} =-K\left[(Dv)^Tm_t + Dm_t\, v_t + m_t \operatorname{div} v_t\right],
\end{equation}
where $D$ denotes the Jacobian matrix. Here $K$ is an inverse operator of $L: V \rightarrow V^*$, which is a positive-definite differential operator that maps a tangent vector $v \in V$ into the dual space $m \in V^*$. This paper employs a commonly used Laplacian operator $L = (-\alpha \Delta + e)^c$, where $\alpha$ is a positive weight parameter, $e$ denotes an identity matrix, and $c$ is a smoothness parameters. A larger value of $\alpha$ and $c$ indicates more smoothness.

A recent model FLASH has later developed a Fourier variant of the EPDiff and demonstrated that the numerical solution to the original EPDiff can be efficiently computed in a low-dimensional bandlimited space with dramatically reduced computational cost~\cite{zhang2015fast,zhang2017frequency}. Let $\widetilde{\Diff}(\Omega)$ and $\tilde{V}$ denote the space of Fourier representations of diffeomorphisms
and velocity fields respectively. The equation of EPDiff is reformulated in a bandlimited space as
\begin{align}\label{eq:epdiffleft}
    \frac{\partial \tilde{v}_t}{\partial t} =-\tilde{K}\left[(\tilde{\mathcal{D}} \tilde{v}_t)^T \star \tilde{m}_t + \tilde{\nabla} \cdot (\tilde{m}_t \otimes \tilde{v}_t) \right],
\end{align}
where $\tilde{v}_t$ and $\tilde{m}_t$ are the Fourier transforms of velocity fields $v_t$ and momentum $m_t$ respectively. The $\tilde{\mathcal{D}}$ represents Fourier frequencies of a Jacobian matrix $D$ with central difference approximation, $\star$ is the truncated matrix-vector field auto-correlation with zero-padding, and $\otimes$ denotes a tensor product. The Fourier coefficients of a laplacian operator $L$ is $\tilde{{L}}(\xi_1 , \ldots, \xi_d) = \left(-2 \alpha \sum_{j = 1}^d \left(\cos (2\pi \xi_j) - 1 \right) + 1\right)^c$, where $(\xi_1 , \ldots, \xi_d)$ is a d-dimensional frequency vector.

\paragraph{\bf LDDMM with geodesic shooting.} Given a source image $S$ and a target image $T$ defined on a $d$-dimensional torus domain $\Omega = \mathbb{R}^d / \mathbb{Z}^d$ ($S(x), T(x):\Omega \rightarrow \mathbb{R}$). 
The space of diffeomorphisms is denoted by $\Diff(\Omega)$.
The problem of diffeomorphic image registration is to find the shortest path, i.e., geodesic, to generate time-varying diffeomorphisms $\{\phi_t\}: t \in [0,\tau] $, such that a deformed source image by the smooth mapping $\phi_\tau$, noted as $S(\phi_\tau)$, is similar to $T$. By parameterizing the transformation $\phi_\tau$ with an initial velocity field $\tilde{v}_0$, we write the optimization of diffeomorphic registration in the setting of LDDMM with geodesic shooting in a low-dimensional bandlimited space as
\begin{align}\label{eq:FEV0}
E(\tilde{v}_0) = \lambda \, \text{Dist}(S(\phi_\tau), T ) + \frac{1}{2}\| \tilde{v}_0 \|_{\tilde{V}}^2 \, \, \text{s.t. Eq.}~\eqref{eq:phi_v} \, \text{and} \, ~\eqref{eq:epdiffleft}
\end{align}
Here, Dist(·,·) is a distance function that measures the dissimilarity between images and $\lambda$ is a positive weighting parameter. The commonly used distance functions include the sum-of-squared intensity differences ($L_2$-norm)~\cite{beg2005computing}, normalized cross correlation (NCC)~\cite{avants2008symmetric}, and mutual information (MI)~\cite{wells1996multi}. In this paper, we will use the sum-of-squared intensity differences. The $\| \cdot \|_V$ represents a Sobolev space that enforces smoothness of the velocity fields. 

Despite a dramatic speed up of LDDMM with the reduced cost of Fourier EPDiff for shooting~\cite{zhang2015fast,zhang2017frequency}, the numerical solution operator in Eq.~\eqref{eq:epdiffleft} does not scale well due to the introduced correlation and tensor convolutions. 

\section{Our Method: NeurEPDiff}
\label{Sec:archNEPdiff}
We introduce a novel network, NeurEPDiff, that learns the solution operator to EPDiff in a low-dimensional bandlimited space. Inspired by the recent work on neural operators~\cite{kovachki2021neural,li2020fourier}, we propose to develop NeurEPDiff, $\mathcal{G}_\theta$ with parameters $\theta$, as a surrogate model to approximate the solution operator to EPDiff. Once it is trained, our NeurEPDiff can be used to fast predict the original EPDiff with a given initial condition $\tilde{v}_0$ on various resolutions of image grids. 

Given a set of $N$ observations, $\{\tilde{v}_0^n , \bm{\tilde{v}}^n \}_{n=1}^N$, where $\bm{\tilde{v}}^n \triangleq \{ \tilde{v}_1^n, \cdots, \tilde{v}_{\tau}^n\}$ is a time-sequence of numerical solutions to EPDiff. The solution operator can be learned by minimizing the empirical data loss defined as 
\begin{equation}
\label{eq:energy}
   E(\theta) = \frac{1}{N} \sum_{n=1}^N \| \bm{\tilde{v}}^n - \mathcal{G}_\theta(v_0^n) \|_{\tilde{V}}^2  + \beta \cdot  \text{Reg}(\theta),  
\end{equation}
where $\beta$ is a weighting parameter and $\text{Reg}(\cdot)$ regularizes the network parameter $\theta$. 

Following the similar principles in~\cite{li2020fourier}, we formulate $\mathcal{G}_\theta$ as an iterative architecture $\tilde{v}_0 \mapsto \tilde{v}_1 \mapsto \cdots \mapsto \tilde{v}_\tau$, which is a sequence of functions carefully parameterized in the hidden network layers. The input $\tilde{v}_0$ is first lifted to a higher dimensional representation by a complex-valued local transformation $\tilde{P}$, which is usually parameterized by a shallow fully connected neural network. We then develop a key component, called {\em iterative evolution layer}, with operations designed in the bandlimited space to learn the nonlinear function of EPDiff. 

\paragraph*{\bf Iterative evolution layer.} Let $\tilde{z} \in \mathbb{C}^d$ be a $d$-dimensional complex-valued latent feature vector encoded from the initial velocity $\tilde{v}_0$. An iterative evolution layer (with the number of $J$ hidden layers) to update $\tilde{z}_j \mapsto \tilde{z}_{j+1}$, where $j \in \{1, \cdots, J\}$, is defined as the composition of a nonlinear smooth activation function $\sigma$ of a complex-valued linear transform $\tilde{W}_j$ with a global convolutional kernel $\mathcal{\tilde{H}}_j$:   
\begin{align}\label{eq:FreFno}
   \tilde{z}_{j+1} := \sigma (\tilde{W}_j\tilde{z}_j + \mathcal{\tilde{H}}_j* \tilde{z}_{j}).
\end{align}
The $*$ represents a complex convolution, which is efficiently computed by multiplying two signals in a real space. Similar to~\cite{scardapane2018complex,Wang_2020_CVPR}, we employ a complex-valued Gaussian error linear unit (CGeLU) that applies real-valued activation functions separately to the real and imaginary parts. With a smoothing operator $\tilde{K}$ to ensure the smoothness of the output signal, we have $\sigma := \tilde{K} \left(\text{GeLU} (\mathscr{R} (\cdot)) + i \text{GeLU} (\mathscr{I} (\cdot))\right)$, where $\mathscr{R} (\cdot)$ denotes the real part of a complex-valued vector, and $\mathscr{I} (\cdot)$ denotes an imaginary part.   

We are now ready to introduce a formal definition of the NeurEPDiff $\mathcal{G}_\theta$ developed in a bandlimited space as follows. 
\begin{definition}[Fourier NeurEPDiff]
The neural operator $\mathcal{G}_\theta$ is defined as a composition of a complex-valued encoder $\tilde{P}$ (a decoder $\tilde{Q}$) (a.k.a. local linear transformations) to project the lower dimension function into higher dimensional space and vice versa, and a nonlinear smooth activation function $\sigma$ of a local linear transform $\tilde{W}$ with a non-local convolution kernel $\mathcal{\tilde{H}}$ parameterized by network parameters, i.e.,   
\begin{equation}
\mathcal{G}_\theta \coloneqq \tilde{Q} \circ \sigma(\tilde{W}_J, \mathcal{\tilde{H}}_J) \circ \cdots \circ \sigma(\tilde{W}_1, \mathcal{\tilde{H}}_1) \circ \tilde{P}.
\label{eq:FneurEPDiff}
\end{equation}
The $\circ$ denotes function composition, and the parameter $\theta$ includes all parameters from $\{\tilde{P}, \tilde{Q}, \tilde{W}, \mathcal{\tilde{H}}\}$. 
\end{definition}

\begin{figure*}[h]
\centering
 \includegraphics[width=1.0\textwidth] {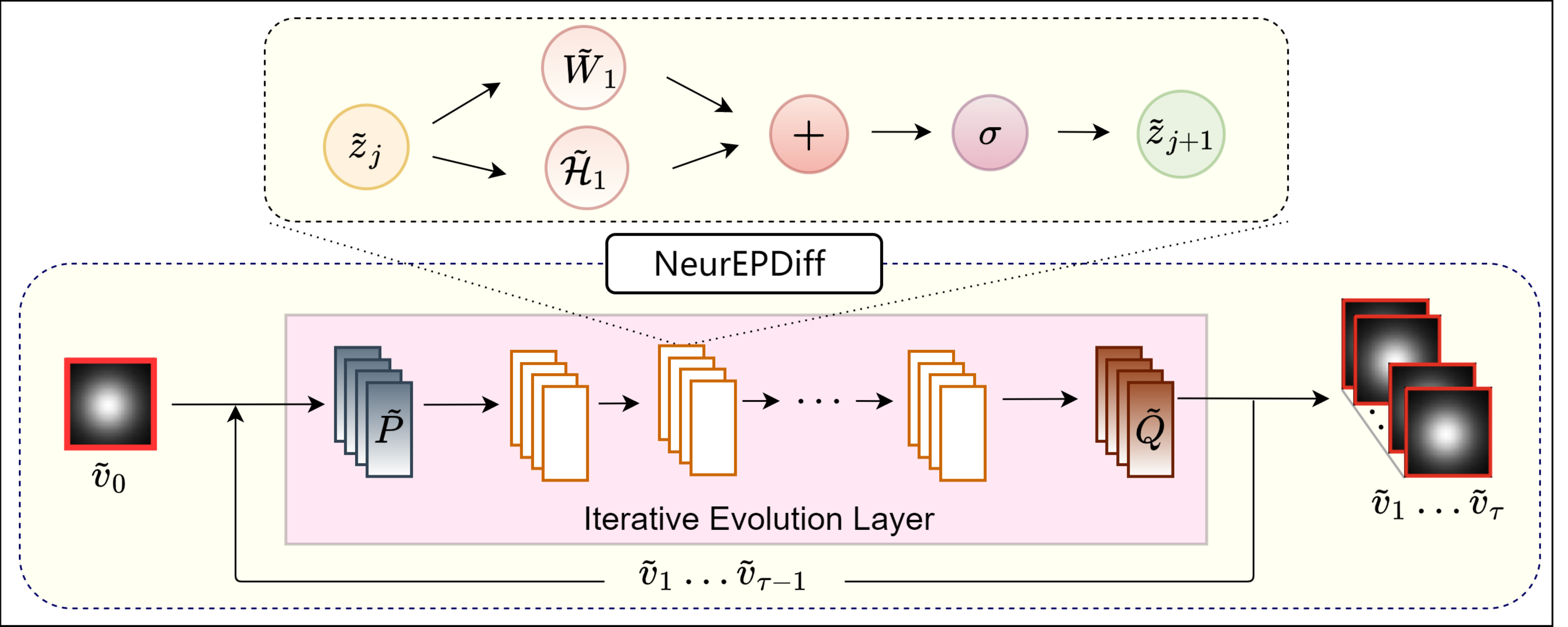}
    %\vspace{-0.4cm}
     \caption{An overview of our proposed NeurEPDiff. Given an initial input $\tilde{v}_0$, the iterative evolution layer updates the time sequence of $\{\tilde{v}_1, \cdots, \tilde{v}_{\tau}\}$.}
\label{fig:architecture_NEpdiff}
\end{figure*}

The network architecture of our proposed NeurEPDiff is shown in Fig.~\ref{fig:architecture_NEpdiff}.   

\subsection{NeurEPDiff For Diffeomorphic Image Registration}
\label{sec:registration}
This section introduces a new diffeomorphic image registration algorithm with our developed NeurEPDiff. We will first train a neural operator $\mathcal{G}_\theta$ offline and then use it as predictive model in each iteration of the optimization of LDDMM (Eq.~\eqref{eq:FEV0}), where the numerical solution of EPDiff is required. In this paper, we use a sum-of-squares-distance (SSD) metric for image dissimilarity. The optimization 
\begin{align}
\label{eq:neurEnergy}
E(\tilde{v}_0) = \frac{1}{2\lambda^{2}}\, \|(S(\phi_{\tau=1}) - T \|_2^2 + \frac{1}{2}\| \tilde{v}_0 \|_{\tilde{V}}^2, \, \, \text{s.t. Eq.}~\eqref{eq:phi_v} \, \& \, \mathcal{G}_\theta(\tilde{v}_0).
\end{align}
The final transformation $\phi_{\tau=1}$ is computed by Eq.~\eqref{eq:phi_v} after predicting $\{\tilde{v}_t\}$ from a well-trained network $\mathcal{G}_\theta$.

\paragraph*{\bf Inference.} Similar to~\cite{zhang2017frequency,vialard2012}, we employ a gradient decent algorithm to minimize the problem~\eqref{eq:neurEnergy}. The three main steps of our inference are described below:
\begin{enumerate}[label=(\roman*)]
    \item Forward geodesic shooting by NeurEPDiff. Given an initialized velocity field $\tilde{v}_0$, we compute the transformation $\phi_\tau$ by first predicting $\{\tilde{v}_t\}$ and then solving Eq.~\eqref{eq:phi_v}.
    %\begin{equation} \label{eq:vupdate}
     %    \{ \tilde{v}_t = g_\theta((\tilde{v}_{t-1})^{pre}) \}_{t= 1 }^{T}.  
   %\end{equation}
    \item Compute the gradient $\nabla_{\tilde{v}_\tau} E$ at the ending time point $t=\tau$ as
    \begin{equation}\label{eq:GradV1}
   \nabla_{\tilde{v}_\tau} E = \tilde{K} \mathcal{F} \left( \frac{1}{\lambda^2} (S(\phi_\tau) - T) \cdot \nabla (S(\phi_\tau))  \right).
    \end{equation} 
    Recall that $\mathcal{F}$ denotes a Fourier transformation.
   \item Backward integration to bring the gradient $\nabla_{\tilde{v}_\tau} E$ back to $t=0$ by integrating the reduced adjoint Jacobi field equations defined in~\cite{zhang2015fast}. 
\end{enumerate}

\begin{figure*}[h]
\centering
 \includegraphics[width=0.95\textwidth] {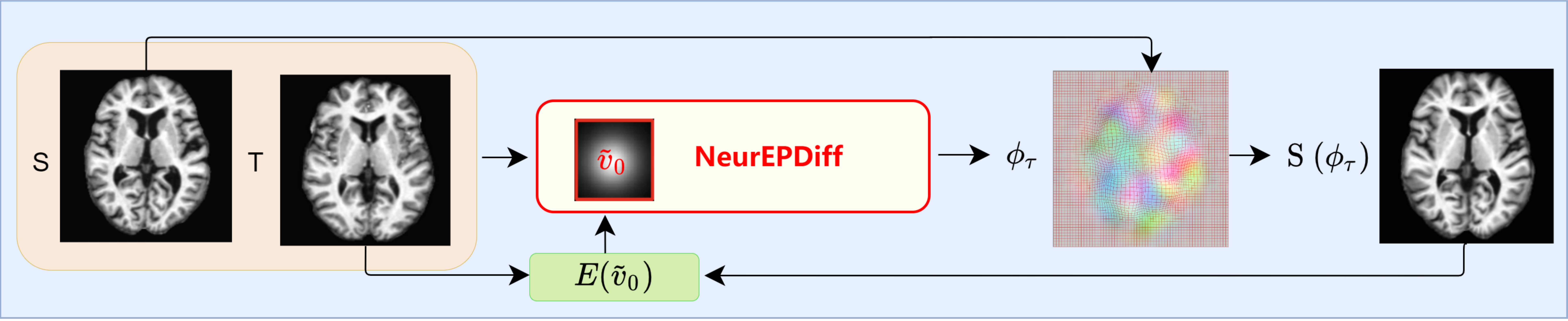}
     \caption{An illustration of our optimization-based diffeomorphic registration with NeurEPDiff.}
      \vspace{-0.4cm}
\label{fig:architecture_Regis}
\end{figure*}

Fig.~\ref{fig:architecture_Regis} visualize the process to optimize image registration with NeurEPDiff. 
%We also summarize the optimization steps in Alg.~\ref{alg:hab}. 

\section{Experimental Evaluation}
To evaluate the effectiveness of our proposed model, NeurEPDiff, for diffeomoprhic image registration, we compare its performance with the original LDDMM with geodesic shooting~\cite{vialard2012} and a fast variant of LDDMM (FLASH)~\cite{zhang2017frequency} on both 2D synthetic data and 3D real brain MR images with different levels of resolution. 

The training of NeurEPDiff in all our experiments is implemented on Nvidia GeForce GTX 1080Ti GPUs. We use ADAM optimizer~\cite{kingma2014adam} with the batch size as $20$, learning rate of $1e^{-3}$, and weight decay as $1e^{-4}$, and the number of epochs as $500$. For a fair comparison with the baseline algorithms that are optimized by gradient decent on CPU, we also test our model on CPU once it is trained.  
%{\color{red} We adopt a learning rate updating strategy of decaying it by $20\%$ every $100$ epochs.} 

We evaluate the registration accuracy by performing registration-based segmentation and examine the resulting segmentation accuracy and runtime. To evaluate volume overlap between the propagated segmentation $A$ and the manual segmentation $B$ for each structure, we compute the dice similarity coefficient DSC$(A, B) = 2(|A| \cap |B|)/(|A| + |B|)$, where $\cap$ denotes an intersection of two regions~\cite{dice1945measures}.

\subsection{Data}
\paragraph*{\bf 2D synthetic data.} We first simulate $2200$ “bull-eye” synthetic data (as shown in Fig~\ref{fig:synthetic}) with the resolution $192^2$ by manipulating the
width $a$ and height $b$ of an ellipse, formulated as
$\frac{(x-96)^2}{a^2} + \frac{(y-96)^2}{b^2} = 1$. We draw the parameters $a, b$
randomly from a Gaussian distribution $\mathcal{N}(40, {4.2}^2)$ for the outer
ellipse, and $a, b \sim \mathcal{N}(17, {1.6}^{2})$ for the inner ellipse. We then carefully down-sample the images to different resolutions of $64^3$ and $128^3$. 
The dataset is split into training ($2000$ volumes of resolution $64^2$) and testing part ($200$ volumes of resolution $64^2$, $128^2$, and $192^2$ respectively). 

\noindent \textbf{3D brain MRI.} We include $2200$ T1-weighted 3D brain MRI scans from Open Access Series of Imaging Studies (OASIS)~\cite{Fotenos1032} in our experiments. We randomly select $2000$ of them and down-sampled to the resolution of $64^3$ for training. The remaining $200$ are re-sampled to the resolution of $64^3$, $128^2$, and $192^2$ for testing separately. The $200$ testing volumes include manually delineated anatomical structures, i.e., segmentation labels, which also have been  carefully down-sampled to the resolution of $64^3$, $128^3$, and $192^3$.
All of the MRI scans have undergone skull-stripping, intensity normalization, bias field correction, and affine alignment. 

\noindent \textbf{Ground truth velocity fields.} We use numerical solutions to Fourier EPDiff in Eq.~\eqref{eq:epdiffleft} (i.e., computed by a Euler integrator) as the ground truth data for network training. More advanced integration methods, such as Runge–Kutta RK4, can be easily applied. In all experiments, we first run pairwise image registration using FLASH algorithm~\cite{zhang2015fast,zhang2017frequency} on randomly selected images with lower resolution of $64^2$ or $64^3$. We then collect all numerical solutions to Fourier EPDiff as ground truth data. We set all integration steps as $10$, the smoothing parameters for the operator $\tilde{K}$ as $\alpha=3.0, c=3.0$, the positive weighting parameter of registration as $\lambda=0.03$, and the dimension of the bandlimited velocity field as $16$.

\subsection{Experiments}
We first test our NeurEPDiff-based registration methods on 2D synthetic data and compare the final estimated transformation fields with the baseline algorithms. To validate whether our model is resolution-invariant, we train NeurEPDiff on images with lower resolution of $64^2$, and then test its prediction accuracy on different levels of higher resolution (i.e., images with $128^2$ and $192^2$). We examine the optimization energy and compare it with all baseline algorithms.  

We further analyze the performance of NeurEPDiff on real 3D brain MRI scans. The convergence graphs of NeurEPDiff-based registration are reported at different image resolutions. To investigate the predictive efficiency of NeurEPDiff in terms of both time-consumption and accuracy, we compare the actual runtime of solving the original EPDiff by our method and all baselines. 

The registration accuracy is validated through a registration-based segmentation. Given a source image, we first run pairwise image registration with all algorithms and propagate segmentation labels of the template image to the number of $100$ target images. We then examine the resulting segmentation accuracy by computing dice scores of the propagated segmentations and manually delineated labels. We report averaged dice scores of all images on four brain structures, including cortex, subcortical-gray-matter (SGM), white-matter (WM), and cerebrospinal fluid (CSF).

\subsection{Results}
\begin{figure*}[h]
\centering
\includegraphics[width=1.0\textwidth] {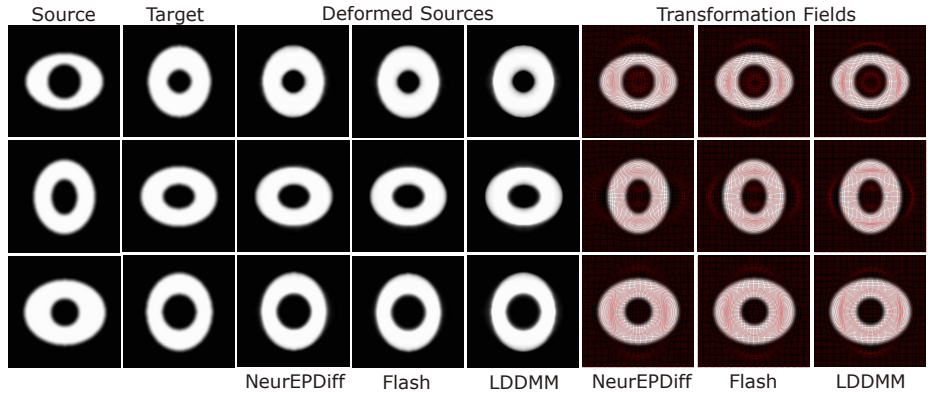}
     \caption{Examples of registration results on 2D synthetic data. Left to right: 2D synthetic source and target images, deformed images, and transformation grids estimated from all algorithms.}
\label{fig:synthetic}
\end{figure*}

Fig.~\ref{fig:synthetic} visualizes examples of registration results on $2$D synthetic images with the size of $192^2$. The comparison is on deformed source images and transformation fields estimated by our NeurEPDiff, FLASH~\cite{zhang2017frequency}, and the original LDDMM with geodesic shooting~\cite{vialard2012}. Note that even though our method is trained on a low image resolution of $64^2$, the predicted results are fairly close to the estimates from registration algorithms of both FLASH and LDDMM on a higher resolution of $192^2$.

\begin{figure*}[t]
\centering
 \includegraphics[width=0.82\textwidth] {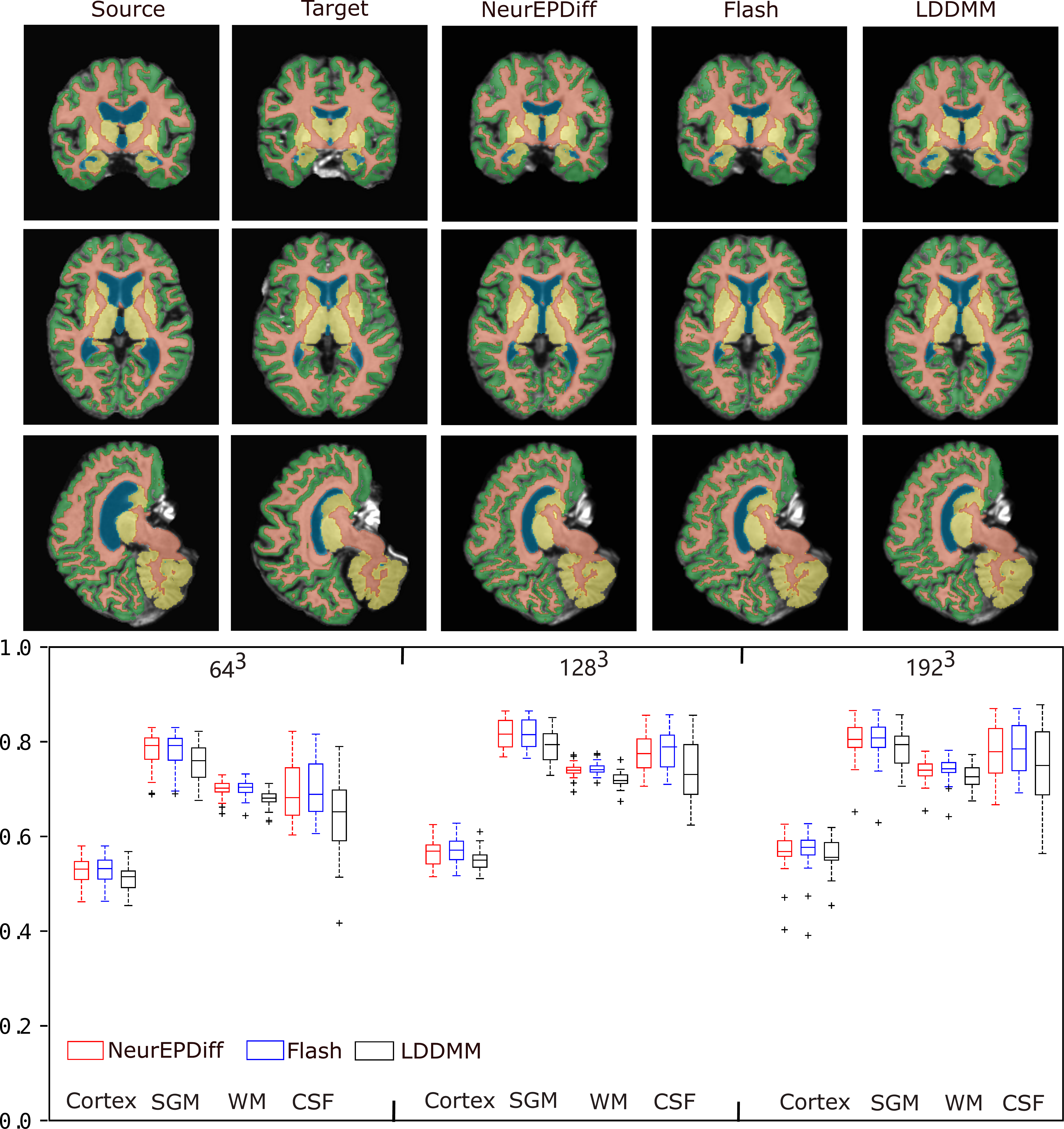}
     \caption{Quantitative result of average dice for all methods on 3D brain MRIs at different scales of resolution: $64^3$, $128^3$, and $192^3$.}
\label{fig:dice}
\end{figure*}

The top panel of Fig.~\ref{fig:dice} displays an example of the comparison between the manually labeled segmentations and propagated segmentations on $3$D brain MRI scans (at the resolution of $192^3$) generated by our algorithm NeurEPDiff and the baseline models. The bottom panel of Fig.~\ref{fig:dice} reports the statistics of average dice scores over four brain structures from $100$ registration pairs at different scales of image resolution. It shows that our method NeurEPDiff produces comparable dice scores. More importantly, the pre-trained NeurEPDiff at a lower resolution of $64^3$ also achieves comparable results when tested on higher resolutions.   

The left panel of Tab.~\ref{tab:runtime} compares the runtime of solving the EPDiff equation with all methods on a CPU machine. It indicates that our method gains orders of magnitude speed compared to other baselines, while maintaining a comparable registration accuracy evaluated through the averaged dice (shown on the the right panel of Tab.~\ref{tab:runtime}). Overall, NeurEPDiff is approximately more than $10$ times faster than FLASH and $100$ times faster than LDDMM with geodesic shooting.
\begin{table*}[!ht]
%\vspace{2mm}
 \begin{center}
     \begin{minipage}{0.5\linewidth}
     \centering
     \setlength{\tabcolsep}{2mm}{
    \begin{tabular}{ccccc}
         \hline
         \multicolumn{2}{c}{\multirow{2}{*}{Methods / Time}} & \multirow{2}{*}{$64^3$} & \multirow{2}{*}{$128^3$} & \multirow{2}{*}{$192^3$} \\
                                                       &               &                 \\ 
                                 \hline
        \multicolumn{2}{c}{NeurEPDiff}               & \bf{0.045s}      & \bf{0.12s} & \bf{0.33s}              \\
         \multicolumn{2}{c}{Flash}              & 0.37s          & 1.24s   & 21.8s               \\ 
         \multicolumn{2}{c}{LDDMM}             & 3.85s         & 30.76s     & 103.09s            \\
                                 \hline
     \end{tabular}
        
         }
     \end{minipage}
%     \hfill
    \hspace{0.2cm}
    \begin{minipage}{0.46\linewidth}
     \centering
     \setlength{\tabcolsep}{2mm}{
     
     \begin{tabular}{ccccc}
        \hline
        \multicolumn{2}{c}{\multirow{2}{*}{Methods / Dice}} & \multirow{2}{*}{$64^3$} & \multirow{2}{*}{$128^3$} & \multirow{2}{*}{$192^3$} \\
                                                       &               &                 \\ 
                                  \hline
        \multicolumn{2}{c}{NeurEPDiff}               & 0.866  & 0.883  & 0.878             \\
         \multicolumn{2}{c}{Flash}              & 0.867       & 0.884   & 0.879              \\ 
         \multicolumn{2}{c}{LDDMM}             & 0.858         & 0.874     & 0.876            \\
                                 \hline
     \end{tabular}    
     }
     \end{minipage}   
     \vspace{2mm}
    \caption{Left to right: a runtime on images with different scales of resolutions and comparison of average dice on all brain structures.}\label{tab:runtime}
 \end{center}
     \vspace{-2mm}
 \end{table*}

\begin{figure*}[h]
\centering
 \includegraphics[width=1.0\textwidth] {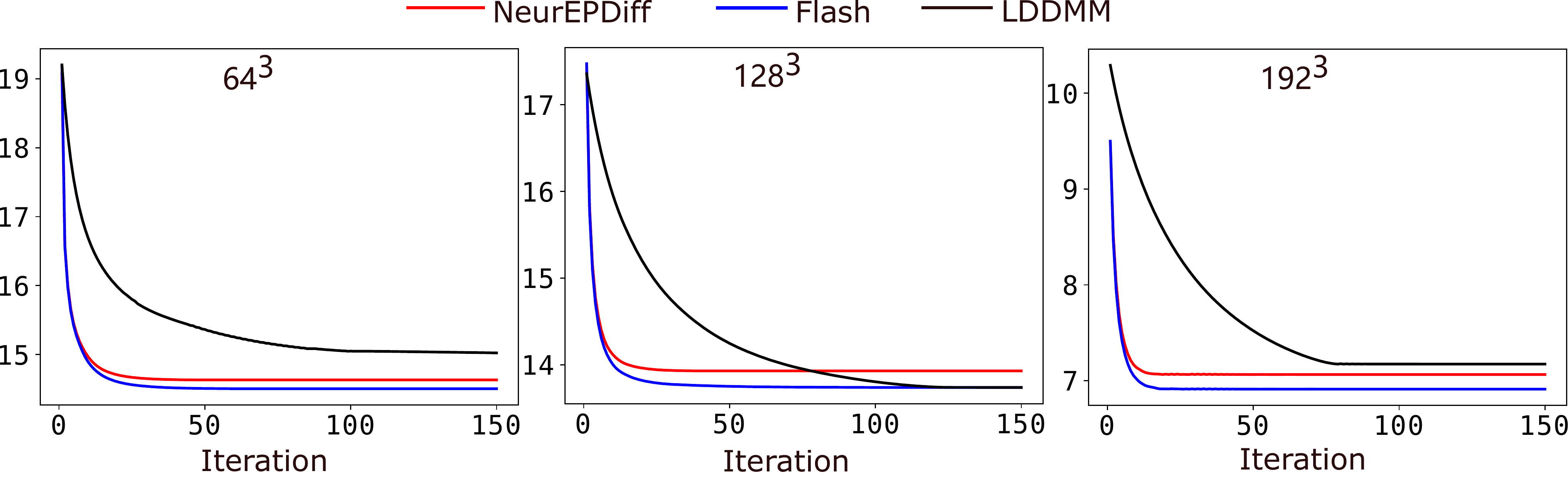}
     \caption{A comparison of convergence graphs of all methods on 3D brain MRIs.}
\label{fig:drawEnergyConvg}
\end{figure*}

Fig.~\ref{fig:drawEnergyConvg} demonstrates the convergence graphs of optimized total energy by our developed NeurEPDiff on different levels of resolution. It shows that our method consistently converges to a better solution than LDDMM at multiple different resolutions. However, it achieves a slightly worse but fairly close solution to FLASH~\cite{zhang2017frequency}. Please refer to Sec.~\ref{sec:discussion} for more detailed discussions. 

\section{Discussion and Conclusion}
\label{sec:discussion}
We presented a novel neural operator, NeurEPDiff, to predict the geodesics of diffeomorphisms in a low-dimensional bandlimited space. Our method is the first to learn a solution operator to the geodesic evolution equation (EPDiff) with significantly improved computational efficiency. More importantly, the prediction capability of a pre-trained NeurEPDiff can be generalized to various resolutions of image grids because of its special property of resolution-invariance. To achieve this, we developed a brand-new network architecture, inspired by recent works on neural operators~\cite{kovachki2021neural,li2020fourier}, to approximate the nonlinear EPDiff equations as time-series functions carefully parameterized in the network hidden layers. A composition of network operators were defined in a complex-valued space to process the bandlimited signals of the velocity fields. 

We validated NeurEPDiff on both simulated 2D and real 3D images. Experimental results show that our method achieves significantly faster speed at different scales of image resolution. Interestingly, we also find out that the NeurEPDiff-based registration algorithm converges to an optimal solution with slightly higher energy compared to FLASH~\cite{zhang2017frequency}. This may be due to a relaxed regularization on the smoothness of the velocity fields parameterized in the network, which often happens in learning-based approaches. A possible solution is to enforce a smooth convolutional kernel $\mathcal{\tilde{H}}$~\cite{feinman2019learning} for each iterative evolution layer of our developed network architecture. This will be further investigated in our future work. Another potential future direction could be applying the network architectures developed in NeurEPDiff to learn another solution operator to predict transformations (Eq.~\eqref{eq:phi_v}), which will further speed up optimization-based and learning-based diffeomorphic registration algorithms~\cite{Wang_2020_CVPR,balakrishnan2019voxelmorph,vialard2012}.

\paragraph*{\bf Acknowledgement.} This work was supported by NSF CAREER Grant 2239977.

{\small
\bibliographystyle{splncs04}
\bibliography{egbib}

\begin{thebibliography}{10}
\providecommand{\url}[1]{\texttt{#1}}
\providecommand{\urlprefix}{URL }
\providecommand{\doi}[1]{https://doi.org/#1}

\bibitem{arnold1966}
Arnold, V.: Sur la g{\'e}om{\'e}trie diff{\'e}rentielle des groupes de lie de
  dimension infinie et ses applications {\`a} l'hydrodynamique des fluides
  parfaits. In: Annales de l'institut Fourier. vol.~16, pp. 319--361 (1966)

\bibitem{arsigny2006log}
Arsigny, V., Commowick, O., Pennec, X., Ayache, N.: A log-euclidean framework
  for statistics on diffeomorphisms. In: International Conference on Medical
  Image Computing and Computer-Assisted Intervention. pp. 924--931. Springer
  (2006)

\bibitem{ashburner2007fast}
Ashburner, J.: A fast diffeomorphic image registration algorithm. Neuroimage
  \textbf{38}(1),  95--113 (2007)

\bibitem{ashburner2005unified}
Ashburner, J., Friston, K.J.: Unified segmentation. Neuroimage  \textbf{26}(3),
   839--851 (2005)

\bibitem{avants2008symmetric}
Avants, B.B., Epstein, C.L., Grossman, M., Gee, J.C.: Symmetric diffeomorphic
  image registration with cross-correlation: evaluating automated labeling of
  elderly and neurodegenerative brain. Medical image analysis  \textbf{12}(1),
  26--41 (2008)

\bibitem{balakrishnan2019voxelmorph}
Balakrishnan, G., Zhao, A., Sabuncu, M.R., Guttag, J., Dalca, A.V.: Voxelmorph:
  a learning framework for deformable medical image registration. IEEE
  transactions on medical imaging  (2019)

\bibitem{beg2005computing}
Beg, M.F., Miller, M.I., Trouv{\'e}, A., Younes, L.: Computing large
  deformation metric mappings via geodesic flows of diffeomorphisms.
  International journal of computer vision  \textbf{61}(2),  139--157 (2005)

\bibitem{cao2015semi}
Cao, T., Singh, N., Jojic, V., Niethammer, M.: Semi-coupled dictionary learning
  for deformation prediction. In: 2015 IEEE 12th International Symposium on
  Biomedical Imaging (ISBI). pp. 691--694. IEEE (2015)

\bibitem{Joshi2004}
Christensen, G.E., Rabbitt, R.D., Miller, M.I.: Deformable templates using
  large deformation kinematics. IEEE Transactions on Image Processing
  \textbf{5}(10),  1435--1447 (1996)

\bibitem{dice1945measures}
Dice, L.R.: Measures of the amount of ecologic association between species.
  Ecology  \textbf{26}(3),  297--302 (1945)

\bibitem{feinman2019learning}
Feinman, R., Lake, B.M.: Learning a smooth kernel regularizer for convolutional
  neural networks. arXiv preprint arXiv:1903.01882  (2019)

\bibitem{Fotenos1032}
Fotenos, A.F., Snyder, A., Girton, L., Morris, J., Buckner, R.: Normative
  estimates of cross-sectional and longitudinal brain volume decline in aging
  and ad. Neurology  \textbf{64}(6),  1032--1039 (2005)

\bibitem{hong2017fast}
Hong, Y., Golland, P., Zhang, M.: Fast geodesic regression for population-based
  image analysis. In: International Conference on Medical Image Computing and
  Computer-Assisted Intervention. pp. 317--325. Springer (2017)

\bibitem{kingma2014adam}
Kingma, D.P., Ba, J.: Adam: A method for stochastic optimization. arXiv
  preprint arXiv:1412.6980  (2014)

\bibitem{kovachki2021neural}
Kovachki, N., Li, Z., Liu, B., Azizzadenesheli, K., Bhattacharya, K., Stuart,
  A., Anandkumar, A.: Neural operator: Learning maps between function spaces.
  arXiv preprint arXiv:2108.08481  (2021)

\bibitem{li2020fourier}
Li, Z., Kovachki, N., Azizzadenesheli, K., Liu, B., Bhattacharya, K., Stuart,
  A., Anandkumar, A.: Fourier neural operator for parametric partial
  differential equations. arXiv preprint arXiv:2010.08895  (2020)

\bibitem{lu2019deeponet}
Lu, L., Jin, P., Karniadakis, G.E.: Deeponet: Learning nonlinear operators for
  identifying differential equations based on the universal approximation
  theorem of operators. arXiv preprint arXiv:1910.03193  (2019)

\bibitem{miller2004computational}
Miller, M.I.: Computational anatomy: shape, growth, and atrophy comparison via
  diffeomorphisms. NeuroImage  \textbf{23},  S19--S33 (2004)

\bibitem{miller2006}
Miller, M.I., Trouv\'{e}, A., Younes, L.: Geodesic shooting for computational
  anatomy. Journal of Mathematical Imaging and Vision  \textbf{24}(2),
  209--228 (2006)

\bibitem{niethammer2011geodesic}
Niethammer, M., Huang, Y., Vialard, F.X.: Geodesic regression for image
  time-series. In: International conference on medical image computing and
  computer-assisted intervention. pp. 655--662. Springer (2011)

\bibitem{o2012unbiased}
O’Donnell, L.J., Wells, W.M., Golby, A.J., Westin, C.F.: Unbiased groupwise
  registration of white matter tractography. In: International Conference on
  Medical Image Computing and Computer-Assisted Intervention. pp. 123--130.
  Springer (2012)

\bibitem{qiu2008parallel}
Qiu, A., Younes, L., Miller, M.I., Csernansky, J.G.: Parallel transport in
  diffeomorphisms distinguishes the time-dependent pattern of hippocampal
  surface deformation due to healthy aging and the dementia of the alzheimer's
  type. NeuroImage  \textbf{40}(1),  68--76 (2008)

\bibitem{Vialard2011DiffeomorphicAE}
Risser, L., Holm, D., Rueckert, D., Vialard, F.X.: Diffeomorphic atlas
  estimation using karcher mean and geodesic shooting on volumetric images. In:
  MIUA (2011)

\bibitem{rohe2017svf}
Roh{\'e}, M.M., Datar, M., Heimann, T., Sermesant, M., Pennec, X.: Svf-net:
  Learning deformable image registration using shape matching. In:
  International Conference on Medical Image Computing and Computer-Assisted
  Intervention. pp. 266--274. Springer (2017)

\bibitem{scardapane2018complex}
Scardapane, S., Van~Vaerenbergh, S., Hussain, A., Uncini, A.: Complex-valued
  neural networks with nonparametric activation functions. IEEE Transactions on
  Emerging Topics in Computational Intelligence  \textbf{4}(2),  140--150
  (2018)

\bibitem{schnabel2001generic}
Schnabel, J.A., Rueckert, D., Quist, M., Blackall, J.M., Castellano-Smith,
  A.D., Hartkens, T., Penney, G.P., Hall, W.A., Liu, H., Truwit, C.L., et~al.:
  A generic framework for non-rigid registration based on non-uniform
  multi-level free-form deformations. In: International Conference on Medical
  Image Computing and Computer-Assisted Intervention. pp. 573--581. Springer
  (2001)

\bibitem{shi2012registration}
Shi, W., Zhuang, X., Pizarro, L., Bai, W., Wang, H., Tung, K.P., Edwards, P.,
  Rueckert, D.: Registration using sparse free-form deformations. In:
  International Conference on Medical Image Computing and Computer-Assisted
  Intervention. pp. 659--666. Springer (2012)

\bibitem{singh2013hierarchical}
Singh, N., Hinkle, J., Joshi, S., Fletcher, P.T.: A hierarchical geodesic model
  for diffeomorphic longitudinal shape analysis. In: International Conference
  on Information Processing in Medical Imaging. pp. 560--571. Springer (2013)

\bibitem{vialard2012}
Vialard, F.X., Risser, L., Rueckert, D., Cotter, C.J.: Diffeomorphic 3d image
  registration via geodesic shooting using an efficient adjoint calculation.
  International Journal of Computer Vision  \textbf{97}(2),  229--241 (2012)

\bibitem{Wang_2020_CVPR}
Wang, J., Zhang, M.: Deepflash: An efficient network for learning-based medical
  image registration. In: Proceedings of the IEEE/CVF Conference on Computer
  Vision and Pattern Recognition (CVPR) (June 2020)

\bibitem{wang2022geo}
Wang, J., Zhang, M.: Geo-sic: Learning deformable geometric shapes in deep
  image classifiers. The Conference on Neural Information Processing Systems
  (2022)

\bibitem{wells1996multi}
Wells, W., Viola, P., Atsumi, H., Nakajima, S., Kikinis, R.: Multi-modal volume
  registration by maximization of mutual information. Medical image analysis
  (1996)

\bibitem{yang2017quicksilver}
Yang, X., Kwitt, R., Styner, M., Niethammer, M.: Quicksilver: Fast predictive
  image registration--a deep learning approach. NeuroImage  \textbf{158},
  378--396 (2017)

\bibitem{younes2009evolutions}
Younes, L., Arrate, F., Miller, M.I.: Evolutions equations in computational
  anatomy. NeuroImage  \textbf{45}(1),  S40--S50 (2009)

\bibitem{zhang2015fast}
Zhang, M., Fletcher, P.T.: Finite-dimensional lie algebras for fast
  diffeomorphic image registration. In: International Conference on Information
  Processing in Medical Imaging. pp. 249--260. Springer (2015)

\bibitem{zhang2017frequency}
Zhang, M., Liao, R., Dalca, A.V., Turk, E.A., Luo, J., Grant, P.E., Golland,
  P.: Frequency diffeomorphisms for efficient image registration. In:
  International Conference on Information Processing in Medical Imaging. pp.
  559--570. Springer (2017)

\bibitem{zhang2016low}
Zhang, M., Wells, W.M., Golland, P.: Low-dimensional statistics of anatomical
  variability via compact representation of image deformations. In:
  International conference on medical image computing and computer-assisted
  intervention. pp. 166--173. Springer (2016)

\end{thebibliography}
}

\end{document}